%% file: main.tex
\documentclass[10pt,twocolumn,letterpaper]{article}

\usepackage[pagenumbers]{wacv} 

\input{preamble}

\definecolor{wacvblue}{rgb}{0.21,0.49,0.74}
\usepackage[pagebackref,breaklinks,colorlinks,allcolors=wacvblue,hypertexnames=false]{hyperref}

\def\wacvPaperID{1275} 
\def\confName{WACV}
\def\confYear{2027}

\title{\ftgspp: Secrets of Dynamic Gaussian Splatting and Their Principles}

\author{
Lucas Yunkyu Lee$^{1,2,*}$ \quad
Soonho Kim$^{1,*}$ \quad
Youngwook Kim$^{1}$\\
Sangmin Kim$^{1}$ \quad
Jaesik Park$^{1,\dagger}$\\[4pt]
$^{1}$Seoul National University \quad
$^{2}$POSTECH\\[3pt]
\small$^{*}$Equal contribution \quad
$^{\dagger}$Corresponding author
}

\begin{document}
\maketitle
\input{sec/0_abstract}    
\input{sec/1_intro}
\input{sec/2_related_work}
\input{sec/3_reproduction}
\input{sec/4_analysis}
\input{sec/5_ftgspp}
\input{sec/6_experiments}
\input{sec/7_limitations_conclusion}

{
    \small
    \bibliographystyle{ieeenat_fullname}
    \bibliography{main}
}

\clearpage
\appendix
\setcounter{page}{1}
\maketitlesupplementary
\input{sec/8_supplementary}

\end{document}

%% file: preamble.tex
\usepackage{multirow}
\usepackage{makecell}
\usepackage{wrapfig}
\usepackage{colortbl}
\usepackage{tabularray}
\UseTblrLibrary{booktabs}

\graphicspath{{figures/}}

\newcommand{\ftgspp}{FreeTimeGS\texttt{++}\xspace}
\newcommand{\ftgsours}{FreeTimeGS\textsubscript{ours}\xspace}
\newcommand{\ftgswang}{FreeTimeGS\textsubscript{Wang}\xspace}
\newcommand{\Base}{\textsc{Base}\xspace}
\newcommand{\UFM}{\textsc{UFM}\xspace}
\newcommand{\Gate}{\textsc{Gate}\xspace}
\newcommand{\CC}{\textsc{CC}\xspace}

\definecolor{colorfirst}{rgb}{.866,.945,0.831}
\definecolor{colorsecond}{rgb}{1,0.98,0.83}
\newcommand{\cellfirst}{\cellcolor{colorfirst}}
\newcommand{\cellsecond}{\cellcolor{colorsecond}}

\newcounter{secret}
\newenvironment{secret}
  {\refstepcounter{secret}%
   \trivlist\item \itshape \textbf{Secret \thesecret.}}
  {\endtrivlist}

%% file: sec/0_abstract.tex
\begin{abstract}
  Recent progress in 4D Gaussian Splatting (4DGS) has achieved impressive
  dynamic scene reconstruction results.
  While these methods demonstrate remarkable performance, the specific factors
  behind their gains remain underexplored, making a systematic understanding of
  the underlying principles challenging.
  In this paper, we perform a comprehensive analysis of these hidden factors to
  provide a clearer perspective on the 4DGS framework.
  We first establish a controlled baseline, \ftgsours, by formalizing and
  reproducing the heuristics of the state-of-the-art FreeTimeGS.
  Using this framework, we examine 4DGS along its fundamental axes and identify
  practical secrets, including the emergent temporal partitioning driven by
  Gaussian durations and the decoupling between photometric fidelity and
  motion behavior.
  Based on these insights, we propose \ftgspp, a principled method that
  employs gated marginalization, UFM-guided initialization, and color correction
  to improve stability and reproducibility.
  Our approach yields reproducible results with reduced run-to-run variance.
  Our code is available at \url{https://yklcs.com/ftgspp}.
\end{abstract}

%% file: sec/1_intro.tex
\section{Introduction}
\label{sec:intro}

The field of dynamic scene reconstruction has recently seen remarkable progress,
driven by advances in 3D Gaussian Splatting (3DGS)~\cite{3dgs}.
A natural extension of this paradigm introduces temporal modeling into Gaussian
representations in order to capture dynamic scenes.
Existing approaches to dynamic Gaussian splatting adopt several different
formulations.
Some methods deform persistent 3D Gaussians through deformation
networks~\cite{4dgs-wu,dgd,gaufre,geomdgs,motiongs,swings,modgs,sc-gs},
while others model temporal dynamics directly within Gaussian primitives using
time-dependent parameters or spacetime
features~\cite{4dgs-yang,stg,longvolcap,gaussianflow,freetimegs}.
Despite these structural differences, many methods report similarly strong
photometric reconstruction performance (\cref{tab:overview_4dgs}).

This diversity raises an important question: \emph{Which structural factors truly
drive the success of 4D Gaussian representations?}
Although these methods share the same underlying rendering framework, their
formulations vary widely in how motion is represented, how primitive lifetimes
are controlled, and how density evolves over time.
Understanding which of these design choices are truly responsible for the strong
empirical performance of 4DGS therefore remains an open challenge.

Many high-performing 4DGS models rely on nuanced implementation details and
heuristics that strongly influence the final reconstruction quality, yet
are not always fully documented in the literature.
As a result, reproducing reported results can be difficult, since performance is
often sensitive to specific choices in datasets or even evaluation frames.
We identify these hidden dependencies as underlying \textit{secrets} of 4DGS.
Addressing these subtleties is essential for providing a clearer perspective on
the principles governing these representations and ensuring their reliable
development and adoption.

In this paper, we perform an extensive analysis of the secrets of 4D Gaussian
Splatting.
We decompose the 4DGS framework into its fundamental components, namely
representation, initialization, training, and evaluation, and systematically
investigate each component to isolate its impact.
Among existing methods, FreeTimeGS provides a modular structure while
maintaining state-of-the-art reconstruction quality~\cite{freetimegs}.
We therefore adopt FreeTimeGS as a strong representative
baseline, but not without additional formalization and analysis.
Throughout our investigation, we formalize previously unpublished
or only loosely defined heuristics, establishing a transparent foundation for
our experiments.
In doing so, we uncover the five \textit{secrets} that explain the mechanics
of 4DGS.

\begin{table*}[t]
    \centering
    \caption{
    \centering
        Design characteristics and reconstruction quality of representative dynamic Gaussian splatting methods. \\
        \dag: Gaussian count limited to 500k for fairness.
    }
    \label{tab:overview_4dgs}
    \vspace{-2mm}
    \begin{tblr}[presep=0pt]{
        colspec={llllcc},
        rows={font=\small},
        row{1}={font=\small\bfseries},
        row{Z}={font=\small\bfseries},
        rowsep=1pt,
        stretch=0.8,
    }
        \toprule
        \SetCell[c=4]{c} Design &&&& \SetCell[c=2]{c} Quality (DyNeRF \cite{dynerf}) & \\
        \cmidrule[lr]{1-4}\cmidrule[lr]{5-6}
        
        Method & Representation & Motion & Density Control & PSNR$\uparrow$ & LPIPS$\downarrow$ \\
        \midrule

        SC-GS~\cite{sc-gs} & canonical + deform & MLP &
        control point ADC & 29.42 & 0.100 \\
        
        4DGS-Wu~\cite{4dgs-wu} & canonical + deform & hexplane + MLP & 3DGS ADC & 31.15 & 0.049 \\ 
        
        STGS~\cite{stg} & spacetime features & polynomial &
        guided sampling & 32.05 & 0.044 \\
        
        4DGS-Yang~\cite{4dgs-yang} & 4D primitives &
        4D rotation & spatiotemporal ADC & 32.01 & 0.055\\
        
        \midrule
        
        FreeTimeGS~\cite{freetimegs}\dag & 4D primitives & linear velocity & relocation & 32.97 & 0.043 \\
        
        \ftgspp\!\dag & gated 4D primitives & linear velocity & relocation & 33.40 & 0.033 \\
        \bottomrule
    \end{tblr}
    \vspace{-2.7mm}
\end{table*}

Drawing on these findings, we propose \ftgspp, a method that explicitly
incorporates these principles to enable robust dynamic scene reconstruction.
We introduce \textit{gated marginalization} to formalize temporal partitioning
and employ \textit{UFM-guided} spatiotemporal initialization, built on a
pre-trained Unified Flow \& Matching model (UFM)~\cite{ufm}, to provide a
motion-aware prior.
We further use a training-time affine color correction module
to improve photometric stability and repeatability.
Extensive evaluations show that \ftgspp improves reconstruction quality
while maintaining comparable perceptual metrics and reducing run-to-run variance.

In summary, our contributions are as follows.
\begin{itemize}
    \item \textbf{Systematic analysis of 4DGS secrets:}
    We reveal the underlying principles behind the success of 4DGS and identify
    the structural shortcomings of existing methods.

    \item \textbf{Formalization of implementation heuristics:}
    We provide concrete formulations of underspecified initialization and training
    heuristics, improving the transparency of the 4DGS pipeline.

    \item \textbf{\ftgspp:}
    A principled 4DGS framework that improves reconstruction quality and repeatability.

    \item \textbf{Reproducible implementation:}
    We release our full implementation and formalized heuristics
    as reusable building blocks to support future 4DGS research.
\end{itemize}

%% file: sec/2_related_work.tex
\section{Related Work}
\label{sec:related-work}

\noindent\textbf{Dynamic scene reconstruction.}
Dynamic scene reconstruction has traditionally been studied through implicit
neural representations.
Early dynamic radiance field methods extend NeRF~\cite{nerf} by modeling time
through canonical-space deformations~\cite{d-nerf, nerfies, hypernerf},
scene-flow parameterizations~\cite{nsff}, or temporal latent
codes~\cite{nerfies}.
While these approaches offer high expressiveness, they typically require complex
training schedules and dense sampling, making large-scale or real-time
applications challenging.

The introduction of explicit neural representations significantly improved
efficiency and rendering speed.
3D Gaussian Splatting (3DGS)~\cite{3dgs} proposed a differentiable splatting
method with adaptive densification and pruning, enabling real-time, high-quality
novel view synthesis.
Building upon this foundation, a growing body of work extends Gaussian
primitives into the temporal domain to model dynamic scenes.

One prominent direction maintains canonical 3D Gaussians and learns
time-dependent deformation fields~\cite{4dgs-wu, ed3dgs, swings, gaufre, dgd}.
These approaches employ deformation networks to predict motion offsets for
Gaussian primitives, sometimes augmented with additional representations such as
per-primitive embeddings~\cite{ed3dgs}.
Related efforts also explore distillation-based pipelines that transfer dynamic
features into Gaussian primitives~\cite{dgd}.

Alternatively, several methods embed temporal dynamics directly within Gaussian
primitives.
These include intrinsic 4D parameterizations with time-conditioned
attributes~\cite{4dgs-yang}, spacetime Gaussian feature splatting~\cite{stg},
sparse timestamp
interpolation schemes~\cite{ex4dgs}, and duration-based activation mechanisms
that allow primitives to appear and disappear over time~\cite{freetimegs}.
Although these methods share the same differentiable splatting framework, they
adopt diverse assumptions regarding primitive lifespan, temporal continuity, and
density control.

\vspace{1mm}\noindent\textbf{Motion consistency in dynamic modeling.}
Although dynamic Gaussian splatting primarily focuses on photorealistic
reconstruction, several works introduce structural or motion-related priors to
encourage more coherent dynamic behavior.
Geometry-aware deformation strategies incorporate spatial consistency priors to
preserve local structure during motion~\cite{geomdgs}, while local rigidity or
neighborhood consistency constraints regularize deformation fields and reduce
excessive non-rigid distortion~\cite{surgicalgs, gaussianflow}.
Other methods leverage external motion cues such as optical flow to align
Gaussian deformation with observed image-space motion~\cite{motiongs, mags,
gflow}.
These approaches motivate analyzing motion behavior beyond photometric scores,
without assuming that image fidelity alone validates the underlying 3D
trajectories.

\vspace{1mm}\noindent\textbf{Optimization stability and reproducibility.}
Beyond representation and motion design, dynamic Gaussian splatting heavily
depends on density control mechanisms and training heuristics.
Adaptive densification and pruning introduced in 3DGS~\cite{3dgs} remain
fundamental in dynamic extensions~\cite{4dgs-wu, 4dgs-yang}.
Additional strategies include progressive training and
backtracking~\cite{ex4dgs}, error-guided refinement~\cite{stg}, sliding-window
optimization~\cite{swings}, cross-temporal consistency
modeling~\cite{timeformer}, and context-aware deformation
learning~\cite{coda4dgs}.
These components substantially influence convergence behavior and reconstruction
stability.

Motivated by these observations, we conduct a controlled analysis of
representation, motion modeling, and density control to better understand the
structural behavior of dynamic Gaussian reconstruction.

%% file: sec/3_reproduction.tex
\section{Preliminary}
\label{sec:preliminaries}

To conduct controlled analysis across representation, initialization, training,
and evaluation, we first establish a transparent baseline.
We choose FreeTimeGS~\cite{freetimegs} as our reference because it achieves
the highest reconstruction quality among existing 4DGS methods with a
simple-yet-effective design.
Since the public training code is unavailable, we implement a best-effort
reproduction, denoted as \ftgsours, by formalizing underspecified components of
the original method, hereafter referred to as \ftgswang, into explicit
algorithms and optimization procedures.

This section first summarizes the core formulation of \ftgswang.
We then document the implementation details for our baseline, \ftgsours, and
finally validate that it recovers competitive performance under matched
primitive budgets.

\subsection{Base Representation}
\label{subsec:base-representation}

Each Gaussian primitive carries a spatial mean
$\boldsymbol{\mu}_x$, a temporal center $\mu_t$, a temporal scale $s$, a velocity
$\mathbf{v}$, covariance parameters (scale $S$ and rotation $R$), a base opacity
$o$, and spherical-harmonics (SH) color coefficients $c$.
Following \ftgswang, its spatial mean moves linearly in time,
\begin{equation}
  \boldsymbol{\mu}_x(t) = \boldsymbol{\mu}_x + \mathbf{v}\,(t - \mu_t),
\end{equation}
and its temporal visibility is governed by a Gaussian-shaped activation
\begin{equation}
    a(t) = \exp\!\left[-\frac{1}{2}\left(\frac{t-\mu_t}{s}\right)^2\right],
    \label{eq:temporal-activation}
\end{equation}
which modulates the base opacity $o$ during rendering.

\subsection{Reimplementation Details}
While \ftgswang outlines the core 4D representation, reproducing its performance
requires explicit formalization of initialization, optimization, and relocation
pipelines.
Here, we detail the specific mechanisms implemented in \ftgsours.

\vspace{1mm}\noindent\textbf{Spatiotemporal initialization.}
The point cloud is initialized with RoMa matching~\cite{roma} for geometric
correspondence. To keep the procedure tractable, each view is matched only to its
$k$ nearest cameras, and triangulation is filtered by cheirality and
coordinate-range checks. Temporal density is controlled by a dataset-specific
keyframe stride, set to 10 for DyNeRF and 1 for SelfCap. Initial velocities are
estimated by $k$-nearest-neighbor correspondence between adjacent keyframe point
clouds and then optimized jointly during training.

\vspace{1mm}\noindent\textbf{Relocation and parameter inheritance.}
\ftgswang periodically reuses low-opacity primitives by relocating them to regions
with high sampling scores, but the strategy for re-initializing the relocated
primitive's parameters is left unspecified. Our baseline adopts an MCMC-inspired
cloning strategy~\cite{3dgs-mcmc}: the relocated primitive inherits the parameters
of its sampled target, while opacity and scale are recomputed following the MCMC
update rule. We keep this as an explicit reproduction choice and revisit its
artifact behavior in \cref{subsec:training}.

\vspace{1mm}\noindent\textbf{Optimization and parameterization.}
The temporal scale $s$ and spatial scale $S$ are optimized in log space and the
base opacity $o$ in logit space to enforce their constraints, while the DC spherical-harmonic term
($sh_0$) is initialized from the input RGB and the higher-order terms ($sh_n$)
start at zero. These choices remove avoidable numerical instability and keep the
baseline suitable for controlled ablations.

\vspace{1mm}\noindent\textbf{Evaluation convention.}
We report PSNR, SSIM/DSSIM, and LPIPS~\cite{perceptual_metric}, averaging over
repeated runs where available. Because LPIPS depends on the input-range convention
assumed by the evaluation code, we compute all LPIPS values under a single
3DGS convention, in which rendered and ground-truth images are kept
in the $[0,1]$ range. 

\begin{table}[t]
  \centering
  \caption{Quantitative comparison on the Neural 3D Video (DyNeRF) and
  SelfCap datasets. $\dagger$ indicates results reported with no more than 500k
  primitives in the original paper~\cite{freetimegs}.}
  \label{tab:reproducing}
  \vspace{-2mm}
  \small
  \setlength{\tabcolsep}{4pt}
  \begin{tabular}{llccc}
    \toprule
    Dataset & Method & PSNR $\uparrow$ & $\text{DSSIM}_2$ $\downarrow$ &
    LPIPS $\downarrow$ \\
    \midrule
    DyNeRF & \ftgswang$^{\dagger}$ & 32.97 & 0.014 & 0.043 \\
    DyNeRF & \ftgsours & 32.62 & 0.011 & 0.033 \\
    \midrule
    SelfCap & \ftgswang$^{\dagger}$ & 27.27 & 0.025 & 0.217 \\
    SelfCap & \ftgsours & 26.43 & 0.027 & 0.137 \\
    \bottomrule
  \end{tabular}
\end{table}

\vspace{1mm}\noindent\textbf{Performance validation.} We quantitatively validate our implementation on the
DyNeRF~\cite{dynerf} and SelfCap~\cite{longvolcap} datasets. As summarized in \cref{tab:reproducing}, \ftgsours recovers
competitive performance against the original \ftgswang, with our numbers averaged
over six independent runs. Detailed per-scene results are provided in the supplement.

%% file: sec/4_analysis.tex
\begin{figure*}[t]
    \centering
    \includegraphics[width=0.9\linewidth]{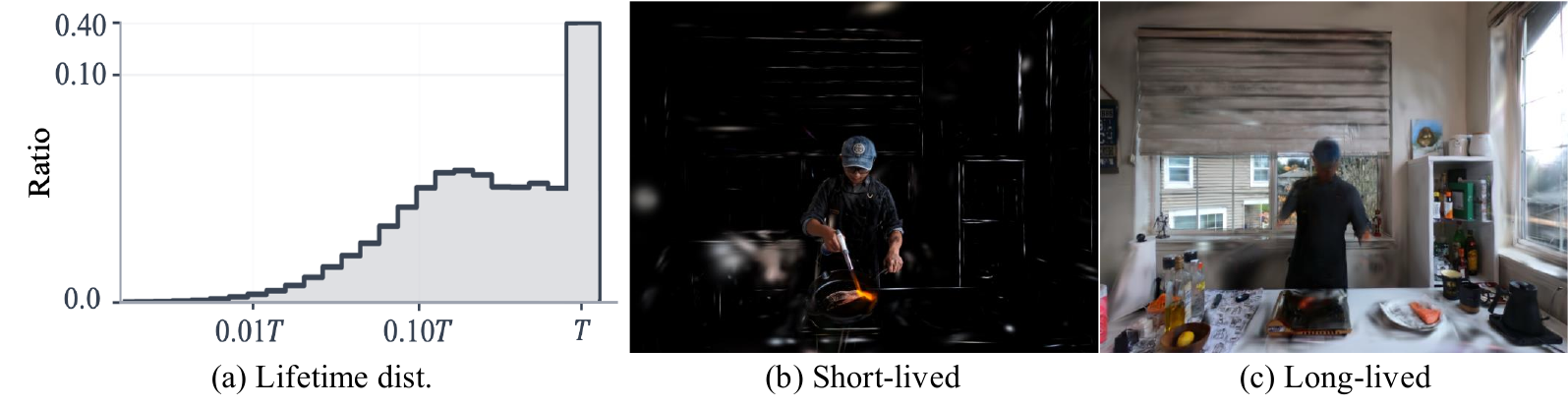}
    \caption{Implicit temporal partitioning by Gaussian lifetime.}
    \label{fig:secret1_partitioning}
\end{figure*}

\section{Secrets Explored}
\label{sec:secrets-explored}

While recent 4DGS methods have demonstrated exceptional performance in
high-fidelity reconstruction, the underlying principles driving these gains are
often entangled with the characteristics of commonly used benchmarks.
Commonly used datasets such as DyNeRF~\cite{dynerf}, SelfCap~\cite{longvolcap},
and other multi-view indoor sequences typically feature a structured environment
comprising static camera arrays, stationary backgrounds, and localized dynamic
subjects.

Such configurations provide a favorable distribution for optimization-based
models to achieve high photometric fidelity (e.g., PSNR), potentially by
over-parameterizing the scene's spatial and temporal priors.
Consequently, the current literature predominantly focuses on optimizing 2D
rendering metrics, often overlooking whether the learned representations
exhibit stable internal motion behavior.

Using \ftgsours as a strong representative baseline, we dissect the 4DGS pipeline
along its fundamental axes, namely representation, initialization, training,
and evaluation, to uncover the hidden \textit{secrets} behind its performance.

\subsection{Representation}
\label{subsec:representation}

\noindent\textbf{Temporal opacity.}
Time-varying opacities constitute one of the two primary mechanisms by which 4D
Gaussians represent dynamic scenes, the other being Gaussian motion.
In duration-based 4D Gaussians such as FreeTimeGS~\cite{freetimegs},
this temporal degree of freedom provides a
simple yet expressive way to control when each primitive contributes to the
rendered sequence.
In this section, we analyze how long each Gaussian remains visible and
how this relates to its functional role in the reconstructed scene.

Each Gaussian is visible only over a limited portion of the
sequence, which we refer to as its \textit{lifetime}. A Gaussian stays visible
longer when it has a larger temporal scale $s$, which widens its active window
around the temporal center, or a higher base opacity $o$, which keeps its
activation above the visibility threshold $\theta$. The lifetime therefore
reflects both factors, not the temporal scale alone.

After convergence, normalizing each lifetime by the
sequence span $T$ reveals a stratified distribution: as shown in
\cref{fig:secret1_partitioning}(a), a substantial fraction of Gaussians stays
visible for nearly the full span $T$, while the rest form a broader distribution
of short-lived primitives. The learned representation therefore does not allocate
temporal support uniformly, but organizes primitives into groups with distinct roles.

\noindent\textbf{Implicit temporal partitioning.}
To examine the functional roles of these groups, we render each
group separately. Rendering only the short-lived Gaussians emphasizes the dynamic
subject (\cref{fig:secret1_partitioning}(b)), whereas the long-lived Gaussians,
whose lifetime approaches the full span $T$, largely recover the static
background (\cref{fig:secret1_partitioning}(c)).

This observation indicates that duration-based 4DGS can develop an implicit temporal
partitioning between transient and persistent scene components.
Even with a simple motion model, temporal activation can allocate
representational capacity to distinct roles in the scene.
Since this partitioning emerges from optimization rather than from an explicit
architectural design, it provides a useful diagnostic for understanding how 4DGS
organizes dynamic content.

\begin{secret}
    Gaussian lifetimes induce an emergent partition between persistent and
    transient scene components.
\end{secret}

\begin{figure}[t]
  \centering
  \includegraphics[width=\linewidth]{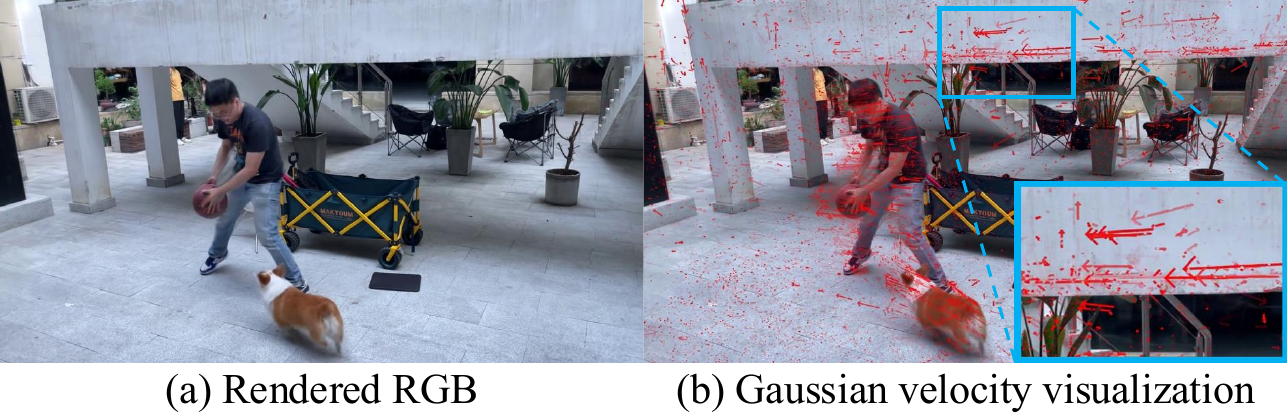}
  \caption{A clean RGB render and its noisy velocity map.}
  \label{fig:secret2_motion}
  \vspace{-5mm}
\end{figure}

\noindent\textbf{Photometric-motion decoupling.}
High-quality photometric reconstruction does not necessarily imply stable
internal motion behavior.
Most 4DGS systems are supervised mainly by rendered image losses, such as
photometric and perceptual reconstruction terms.
These objectives constrain how closely each rendered frame matches the target
image, but not whether a primitive preserves a consistent temporal identity or
trajectory.

This decoupling becomes visible in \ftgsours.
As shown in \cref{fig:secret2_motion}(a,b), the rendered RGB can appear clean
while the velocity map reveals high-frequency noise and motion leakage in
regions that should remain largely static.
Standard rendering metrics can thus hide unstable internal motion behavior.

Such behavior can arise, for example, through
\textit{role-switching}: instead of tracking the same dynamic content with a stable set of primitives along continuous trajectories, the model can change which temporally active primitives explain that content at different times. Photometric consistency is preserved, while the underlying velocity field becomes noisy or temporally fragmented.

\begin{secret}
    High-quality rendering does not necessarily identify a motion-consistent
    dynamic representation.
\end{secret}


\subsection{Initialization}
\label{subsec:initialization}

The high-dimensional nature of 4DGS makes it particularly sensitive to the
quality of the initial point cloud and its temporal distribution.
We analyze how both spatial and temporal density in the initialization phase
influence the final reconstruction fidelity.

\begin{figure}[t]
  \centering
  \includegraphics[width=\linewidth]{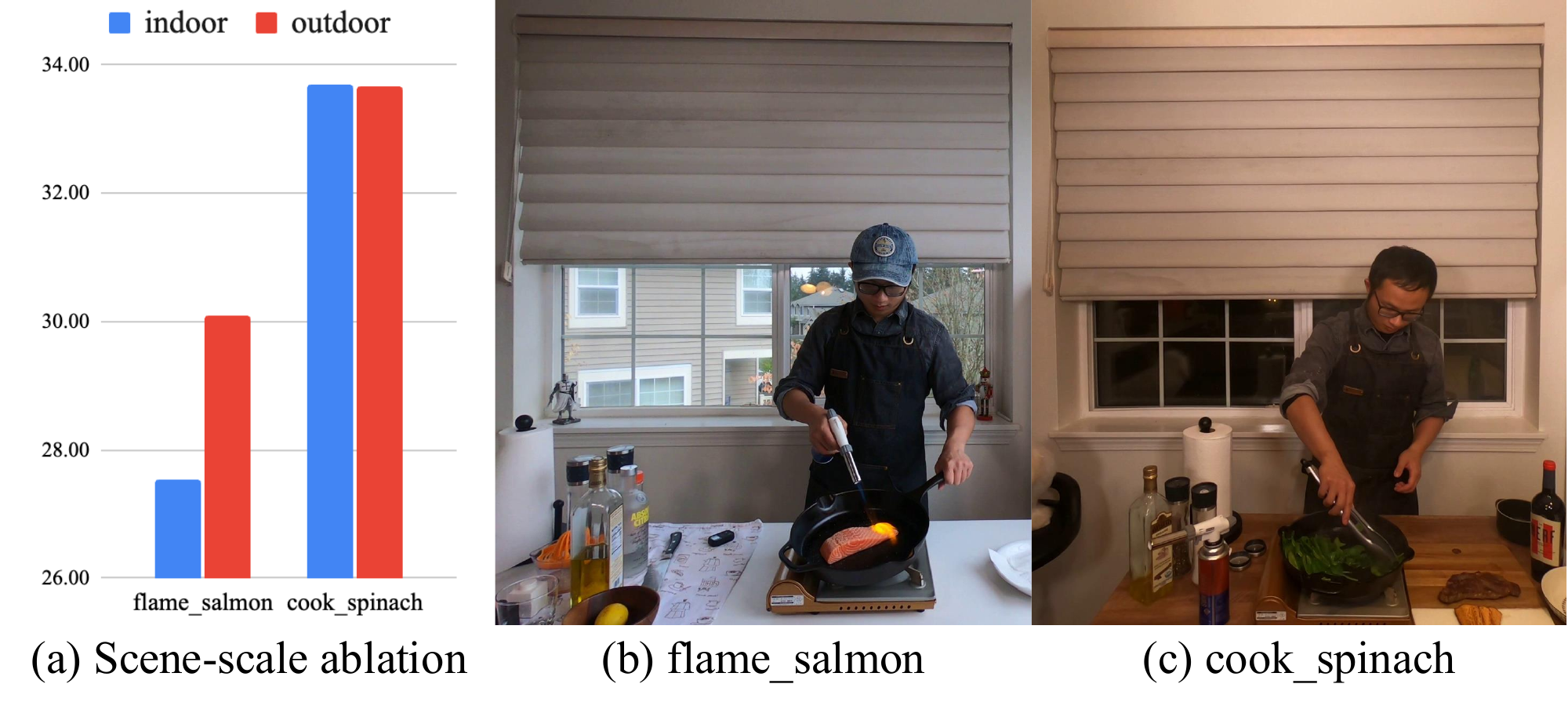}
  \caption{Effect of scene-scale priors in spatial
  initialization.}
  \label{fig:scene-scale-prior}
  \vspace{-5mm}
\end{figure}

\vspace{2mm}\noindent\textbf{Spatial initialization.}
Spatial initialization depends not only on the matching model,
but also on the scene-scale assumption used during point reconstruction.
In particular, the assumed depth range can strongly affect the geometry of the
initial point cloud.
In our reproduced baseline, we use RoMa~\cite{roma} for neural-based dense matching.
RoMa allows us to explicitly choose between indoor and outdoor scene priors,
making it possible to examine how this scene-scale assumption affects
initialization.

This prior matters even for sequences that are nominally indoor.
As shown in \cref{fig:scene-scale-prior}, changing only the scene-scale prior
substantially improves \texttt{flame\_salmon}, where the window exposes outdoor
structures with extended depth variation, while it has little effect on
\texttt{cook\_spinach}, where the window is mostly occluded by a curtain.
Some indoor sequences in DyNeRF and SelfCap contain expansive structures, such as deep corridors or windowed backgrounds.
The depth range of these scenes is closer to an unbounded scene than to
a strictly bounded indoor room.
In such cases, using an outdoor-like prior can produce more reliable geometry
than applying a strictly indoor prior.
This shows that spatial initialization should be aligned with the actual
geometric context of the scene, rather than determined only by the dataset label
or by simply switching to a newer matcher.

\begin{table}[t]
  \centering
  \caption{Temporal stride ablation under a fixed 500k-primitive
  budget. Bold marks the default baseline stride.}
  \vspace{-2mm}
  \label{tab:init_ablation}
  \resizebox{\linewidth}{!}{%
  \begin{tabular}{ccccccc}
    \toprule
    \multicolumn{3}{c}{DyNeRF} & \phantom{a} & \multicolumn{3}{c}{SelfCap} \\
    \cmidrule(lr){1-3}\cmidrule(lr){5-7}
    Stride & PSNR $\uparrow$ & LPIPS$_\mathrm{Alex}\downarrow$ &&
    Stride & PSNR $\uparrow$ & LPIPS$_\mathrm{VGG}\downarrow$ \\
    \midrule
    50 & 32.55 & 0.033 && 10 & 26.22 & 0.143 \\
    20 & 32.56 & 0.032 && 5  & 26.28 & 0.141 \\
    \textbf{10} & \textbf{32.62} & \textbf{0.033} &&
    \textbf{1}  & \textbf{26.43} & \textbf{0.137} \\
    5  & 32.57 & 0.033 && -- & -- & -- \\
    \bottomrule
  \end{tabular}}
  \vspace{-4mm}
\end{table}

\vspace{2mm}\noindent\textbf{Temporal initialization.}
In the temporal dimension, the keyframe stride controls the
temporal resolution of the initialization.
A smaller stride samples the dynamic sequence more densely, while a larger
stride provides a coarser set of initial point clouds.
In our reproduced baseline, the initial velocity is estimated from
correspondences between neighboring keyframe point clouds.
Therefore, the stride determines not only the number of temporal samples, but
also the temporal resolution of the motion cues available at initialization.
We evaluate this effect under a fixed Gaussian budget to separate temporal
sampling from model capacity.

The results in \cref{tab:init_ablation} show that temporal
density is not a monotonic factor.
On DyNeRF, where motion is relatively slow or localized over a longer temporal
span, the default baseline stride of 10 already provides sufficient temporal
coverage.
Reducing the stride further to 5 does not improve the result.
In contrast, SelfCap contains faster human motion over shorter sequences, where
using every frame preserves temporal information that would otherwise be
skipped.

Thus, the results do not support the view that denser
initialization is always better.
Rather, temporal initialization should be matched to the temporal structure of the target sequence.
The useful stride depends on the sequence length, frame rate, and motion complexity.

\begin{secret}
  Initialization quality depends on scene-matched spatial priors
  and motion-matched temporal sampling.
\end{secret}

\subsection{Training}
\label{subsec:training}

\noindent\textbf{Density control.}
The training dynamics of 4DGS are strongly influenced by density
control, which reuses inactive Gaussian primitives by relocating them to more
informative regions.
During training, primitives whose base opacity falls below a visibility threshold
($o \le \theta$) are classified as part of the dead set.
Rather than permanently discarding them, we sample new target locations from the
active population according to primitive opacity and view-space gradient
magnitudes, encouraging relocation toward regions with high rendering
contributions.

However, the effect of relocation is governed not only by where
primitives are moved, but also by which parameters they inherit from the sampled
target primitive.
We compare three inheritance strategies.
\textbf{Partial Copy} copies the target primitive's parameters except opacity and
scale, which are preserved from the inactive primitive.
\textbf{Exact Copy} duplicates all target parameters, including opacity and
scale.
\textbf{MCMC} follows MCMC-based density control~\cite{3dgs-mcmc}, recomputing
opacity and scale to account for the number of cloned primitives.

\begin{figure}[t]
  \centering
  \includegraphics[width=\linewidth]{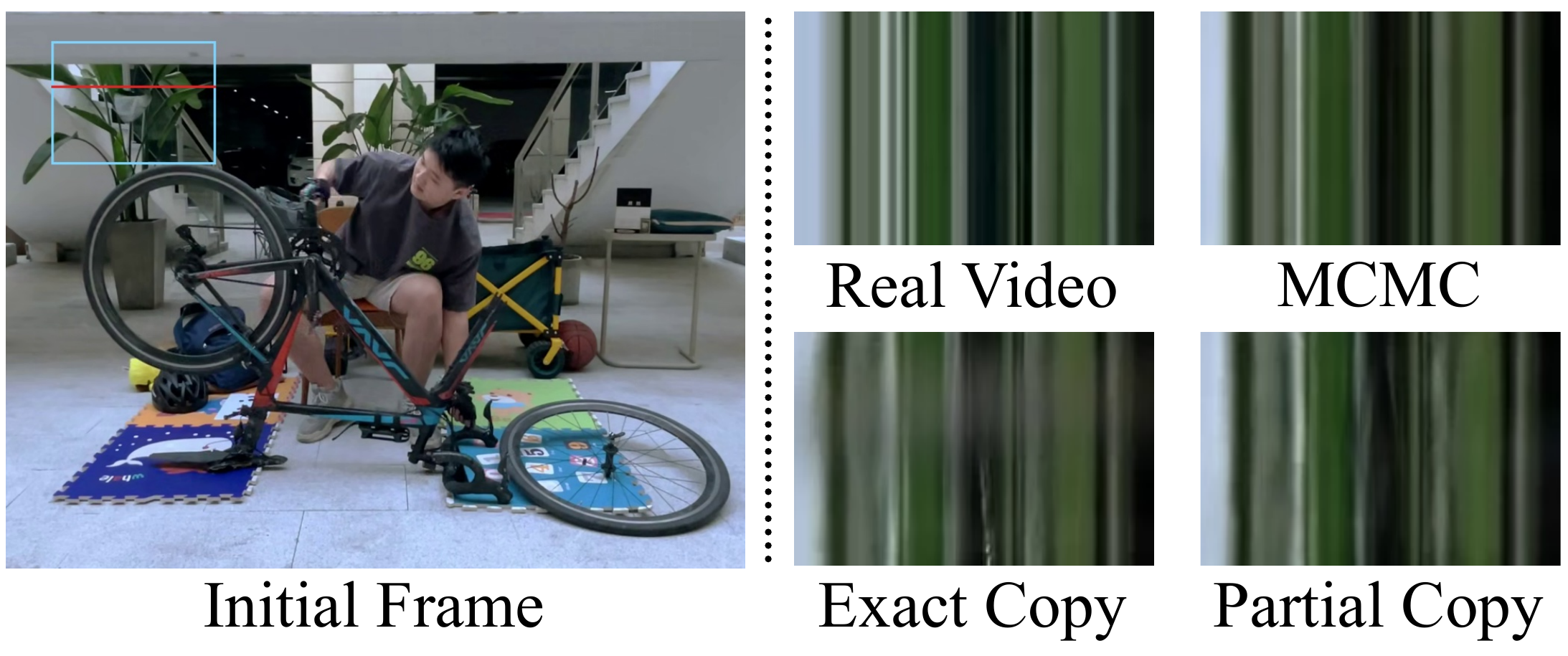}
  \caption{Space-time slices under different relocation strategies.}
  \label{fig:xt-slice}
\end{figure}

\begin{table}[t]
  \centering
  \caption{Impact of relocation strategies on PSNR.}
  \label{tab:density-control}
  \vspace{-2mm}
  \small
  \setlength{\tabcolsep}{5pt}
  \begin{tabular}{lcc}
    \toprule
    Relocation Policy & DyNeRF & SelfCap \\
    \midrule
    MCMC & 32.62 & 26.43 \\
    Exact Copy & \textbf{32.64} & \textbf{26.58} \\
    Partial Copy & 32.55 & 26.09 \\
    \bottomrule
  \end{tabular}
\end{table}

\noindent\textbf{Metric-artifact mismatch.}
As shown in \cref{tab:density-control}, Exact Copy achieves the highest PSNR on both datasets.
However, this metric gain can be misleading.
By fully duplicating high-contribution targets, Exact Copy can accumulate
redundant primitives in the same spatial regions and produce excessive rendering contributions.
In space-time slices, this appears as haze-like wobbling artifacts in static background regions (\cref{fig:xt-slice}).

MCMC does not obtain the highest PSNR, but it provides more
stable qualitative behavior among the tested strategies.
By redistributing opacity and scale during cloning, it suppresses over-amplified
background contributions and yields more stable space-time boundaries.
These results suggest that rendering metrics alone do not fully characterize
relocation strategies, as inherited opacity and scale also shape artifact
accumulation.

\begin{secret}
  Opacity and scale inheritance can determine whether relocation
  stabilizes reconstruction or accumulates background artifacts.
\end{secret}

\subsection{Evaluation}
\label{subsec:evaluation_stability}

Dynamic Gaussian models achieve expressive reconstruction by optimizing many
primitives together with their geometry, opacity, color, temporal activation,
and motion-related parameters.
This flexibility is useful for representing dynamic scenes, but it also creates
a broad optimization space in which different runs can explain the same
frame-wise photometric observations in different ways.
As a result, even under the same training configuration, small stochastic differences
can lead to noticeable run-to-run variation, and a single-run metric may not
fully capture the behavior of the optimization.

\begin{figure}[t]
  \centering
  \includegraphics[width=\linewidth]{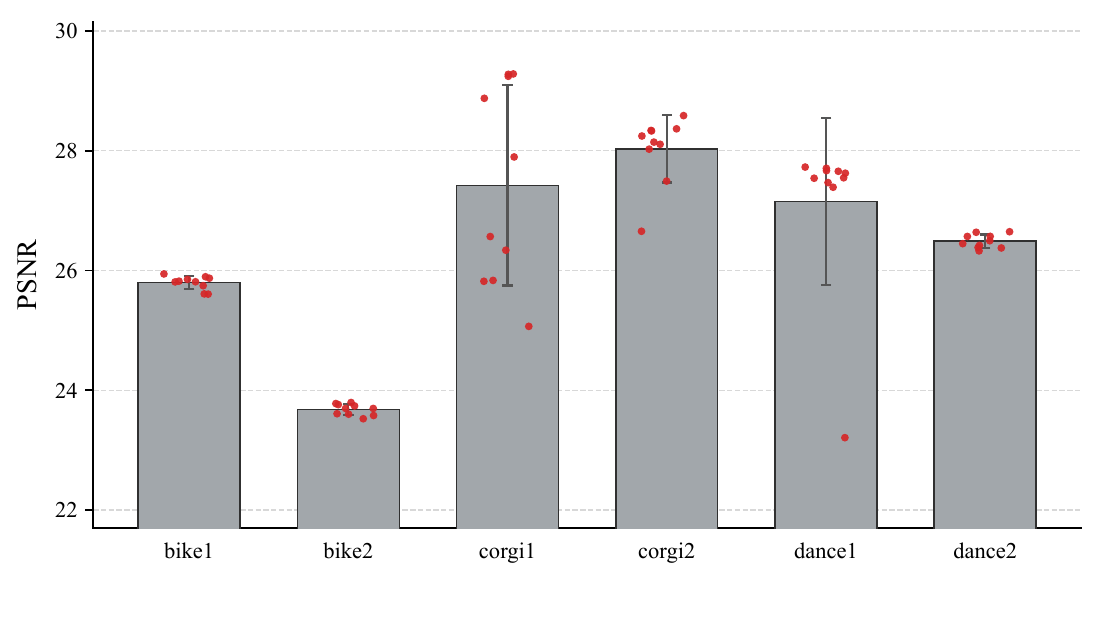}
  \vspace{-2em}
  \caption{Run-to-run PSNR variation on SelfCap. Bars denote
  scene-wise means over 10 runs, and points denote individual runs.}
  \label{fig:selfcap_psnr_stability}
  \vspace{-2mm}
\end{figure}

We visualize this effect on the SelfCap dataset in
\cref{fig:selfcap_psnr_stability}.
All runs use the same training configuration, and the figure summarizes the PSNR
distribution over 10 independent runs for each scene.
For some scenes, the same method and training configuration produce noticeably
different results across runs.

These results indicate that a single reported score may not fully
represent the behavior of a dynamic Gaussian reconstruction pipeline.
This motivates reporting repeated-run behavior when analyzing 4DGS optimization,
especially when differences between methods or design choices are small.

\begin{secret}
  Single-run scores can hide run-to-run variation in dynamic
  Gaussian reconstruction.
\end{secret}

%% file: sec/5_ftgspp.tex
\section{Improved Model: \ftgspp}
\label{sec:improved}

\begin{table*}[t]
  \centering
  \caption{Component-wise ablation of \ftgspp. \CC denotes training-time color correction.}
  \label{tab:component-ablation}
  \label{tab:final-model-improvement}
  \small
  \setlength{\tabcolsep}{5pt}
  \begin{tabular}{llcccccc}
    \toprule
    & & \multicolumn{3}{c}{DyNeRF} & \multicolumn{3}{c}{SelfCap} \\
    \cmidrule(lr){3-5}\cmidrule(lr){6-8}
    Method & Configuration
    & PSNR $\uparrow$ & SSIM$_2 \uparrow$ & LPIPS$_\mathrm{Alex} \downarrow$
    & PSNR $\uparrow$ & SSIM$_2 \uparrow$ & LPIPS$_\mathrm{VGG} \downarrow$ \\
    \midrule
    \ftgsours & \Base
    & 32.62 & 0.978 & 0.033
    & 26.43 & 0.946 & \textbf{0.137} \\
     & \Base + \UFM
    & 32.73 & \textbf{0.979} & \textbf{0.031}
    & 26.25 & 0.945 & 0.138 \\
     & \Base + \UFM + \Gate
    & 32.78 & 0.978 & 0.034
    & 26.41 & 0.945 & 0.139 \\
    \midrule
    \ftgspp & \Base + \UFM + \Gate + \CC
    & \textbf{33.40} & 0.978 & 0.033
    & \textbf{27.12} & \textbf{0.947} & \textbf{0.137} \\
    \bottomrule
  \end{tabular}
\end{table*}

Building on the observations in \cref{sec:secrets-explored}, 
we construct \ftgspp as a compact configuration of analysis-driven refinements.

\subsection{Gated Marginalization}
\label{subsec:improved-gate}

\begin{figure}[t]
  \centering
  \includegraphics[width=0.68\linewidth]{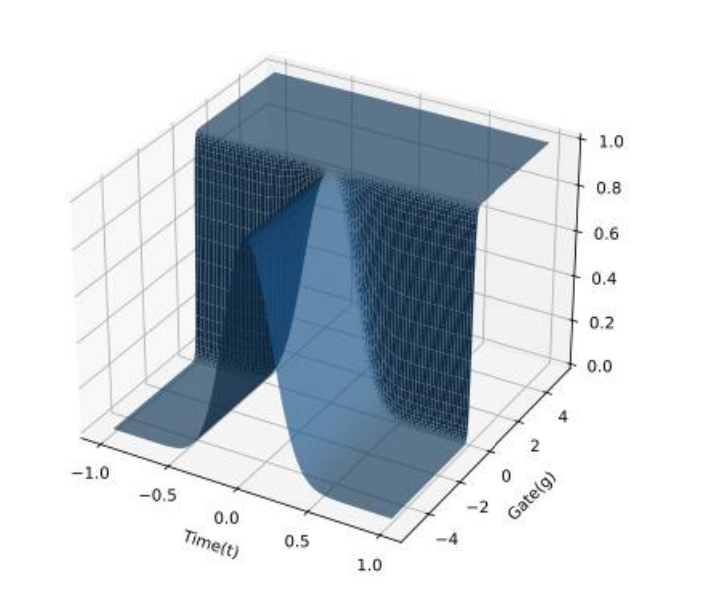}
  \caption{Gated marginalization function $\phi_i(t)$.}
  \label{fig:gated_marginalization}
\end{figure}

The representation analysis in \textbf{\textit{Secret 1}} (\Cref{subsec:representation}) shows that duration-based 4D
Gaussians can implicitly separate persistent and transient components.
Gated marginalization makes this behavior explicit by assigning each Gaussian a
learnable persistence gate.
For Gaussian $i$, we define
\begin{equation}
g_i
=
\sigma(\gamma \tilde{g}_i),
\end{equation}
where $\tilde{g}_i$ is the gate logit and $\gamma$ controls the
transition sharpness.
The gate blends a persistent contribution with the original temporal
activation:
\begin{equation}
\phi_i(t)
=
g_i
+ (1 - g_i)
\exp\left(
-\frac{1}{2}
\left(\frac{t - \mu_{t,i}}{s_i}\right)^2
\right).
\end{equation}
Here, $\mu_{t,i}$ is the temporal center and $s_i=d_i/6$ is the
temporal scale associated with duration $d_i$.
The rendered opacity contribution is
\begin{equation}
\alpha_i(t)=o_i\,\phi_i(t),
\end{equation}
where $o_i$ is the base opacity.
When $g_i$ approaches one, the primitive behaves persistently across time.
When $g_i$ approaches zero, it remains localized around $\mu_{t,i}$.
This continuous parameterization encourages temporal role assignment without
introducing a hard static-dynamic split.

\subsection{UFM-Guided Initialization}
\label{subsec:improved-ufm}

The initialization analysis in \textbf{\textit{Secret 3}} (\Cref{subsec:initialization}) shows that temporal sampling and
motion priors can affect the final reconstruction.
FreeTimeGS-style initialization estimates velocity from nearest-neighbor
correspondences between adjacent keyframe point clouds.
This estimate is simple, but it can be noisy when the initial geometry or
temporal sampling is imperfect.

To provide a more motion-aware starting point, we use priors from
the pre-trained Unified Flow \& Matching model (UFM)~\cite{ufm}.
We project initial 3D points to the image plane, sample multi-view optical
flows, and back-project the flow observations to estimate a UFM-guided velocity
$\mathbf{v}_{i}^{\mathrm{ufm}}$.
When the UFM estimate is valid, it replaces the nearest-neighbor velocity.
Otherwise, we keep the baseline estimate.
\begin{equation}
\mathbf{v}_{i}^{0}
=
m_i \mathbf{v}_{i}^{\mathrm{ufm}}
+ (1 - m_i) \mathbf{v}_{i}^{\mathrm{knn}},
\end{equation}
where $m_i \in \{0,1\}$ is a validity mask based on confidence
and multi-view consistency.
This preserves the baseline initialization in unsupported regions while using
UFM cues where they provide reliable motion information.

\subsection{Training-Time Color Correction}
\label{subsec:cc}

The evaluation analysis in \textbf{\textit{Secret 5}} (\Cref{subsec:evaluation_stability}) suggests the need to consider
run-to-run stability in dynamic Gaussian reconstruction.
In this section, we propose a lightweight affine color correction (CC) module which acts as a training-time stabilizer that absorbs photometric
variation, such as global illumination shifts and camera-dependent color changes.

For training camera $c$, the rendered RGB value is corrected as
\begin{equation}
  \hat{\mathbf{I}}^{\mathrm{cc}}_c
  =
  \hat{\mathbf{I}}(\mathbf{I}_3+\Delta\mathbf{M}_c) + \mathbf{b}_c,
\end{equation}
where $\Delta\mathbf{M}_c\in\mathbb{R}^{3\times3}$ and
$\mathbf{b}_c\in\mathbb{R}^{3}$ define a camera-wise affine color transform.
The correction is regularized toward the identity transform and is used only in the training losses.
Final renderings are produced without applying the CC module.

\begin{table}[t]
  \centering
  \caption{PSNR (mean $\pm$ std) over 10 repeated optimizations
  with fixed initialization.}
  \label{tab:cc-psnr-stability}
  \small
  \setlength{\tabcolsep}{8pt}
  \begin{tabular}{lcc}
    \toprule
    Dataset & Baseline & Baseline + CC \\
    \midrule
    DyNeRF & 32.47 $\pm$ 0.22 & \textbf{32.81 $\pm$ 0.08} \\
    SelfCap & 26.43 $\pm$ 0.36 & \textbf{27.32 $\pm$ 0.09} \\
    \bottomrule
  \end{tabular}
\end{table}

To evaluate the effect of this training-time stabilizer, we fix the initialization and repeat optimization under the same protocol.
This isolates the effect of CC from initialization-induced variation.
As shown in \cref{tab:cc-psnr-stability}, CC reduces
run-to-run variance in the reproduced baseline while also improving mean PSNR.
We hypothesize that CC allows Gaussian primitives to focus on learning geometry and motion, rather than compensating for photometric inconsistencies that may otherwise amplify optimization variance.

\subsection{Training Objective}
\label{subsec:improved-objective}
The full training objective combines the baseline reconstruction
losses with the regularizers introduced above:
\begin{equation}
\mathcal{L}
=
\lambda_1\mathcal{L}_{1}
+ \lambda_{\mathrm{s}}\mathcal{L}_{\mathrm{ssim}}
+ \mathcal{L}_{\mathrm{reg}}
+ \mathcal{L}_{\mathrm{gate}}
+ \mathcal{L}_{\mathrm{cc}}.
\end{equation}
$\mathcal{L}_{\mathrm{reg}}$ denotes the temporal-opacity
regularization inherited from \ftgswang,
$\mathcal{L}_{\mathrm{gate}}$ regularizes the gate parameters, and
$\mathcal{L}_{\mathrm{cc}}$ keeps the affine color correction close to the
identity transform.
Together, gated marginalization, UFM-guided initialization, and training-time
color correction form the main \ftgspp configuration evaluated in
\cref{sec:experiments}.

%% file: sec/6_experiments.tex
\section{Experiments}
\label{sec:experiments}

We conduct experiments on four datasets, Neural 3D
Video (DyNeRF)~\cite{dynerf}, SelfCap~\cite{longvolcap},
ENeRF-Outdoor~\cite{lin2022enerf}, and Google Immersive~\cite{broxton2020immersive}.
DyNeRF consists of six scenes, each captured by 19--21 synchronized cameras.
SelfCap contains six challenging scenes with fast and dynamic motions, while
ENeRF-Outdoor and Google Immersive include dynamic outdoor activities.
We report PSNR, SSIM~\cite{ssim}, and LPIPS~\cite{perceptual_metric} to assess the rendering
quality of our method and the reproduced baseline.

\vspace{1mm}\noindent\textbf{Component-wise ablation.}
\label{subsec:component-ablation}
\Cref{tab:component-ablation} reports the ablation study for
\ftgspp.
UFM-guided initialization provides motion-aware initialization cues, gated
marginalization introduces an explicit persistent/transient temporal role, and
color correction contributes the largest PSNR gain in the final configuration.
The final row corresponds to the main \ftgspp configuration used in the following
experiments.
Overall, \ftgspp yields consistent improvements over our reproduced baseline
across both DyNeRF and SelfCap datasets.

\begin{figure}[t]
  \centering
  \includegraphics[width=\linewidth]{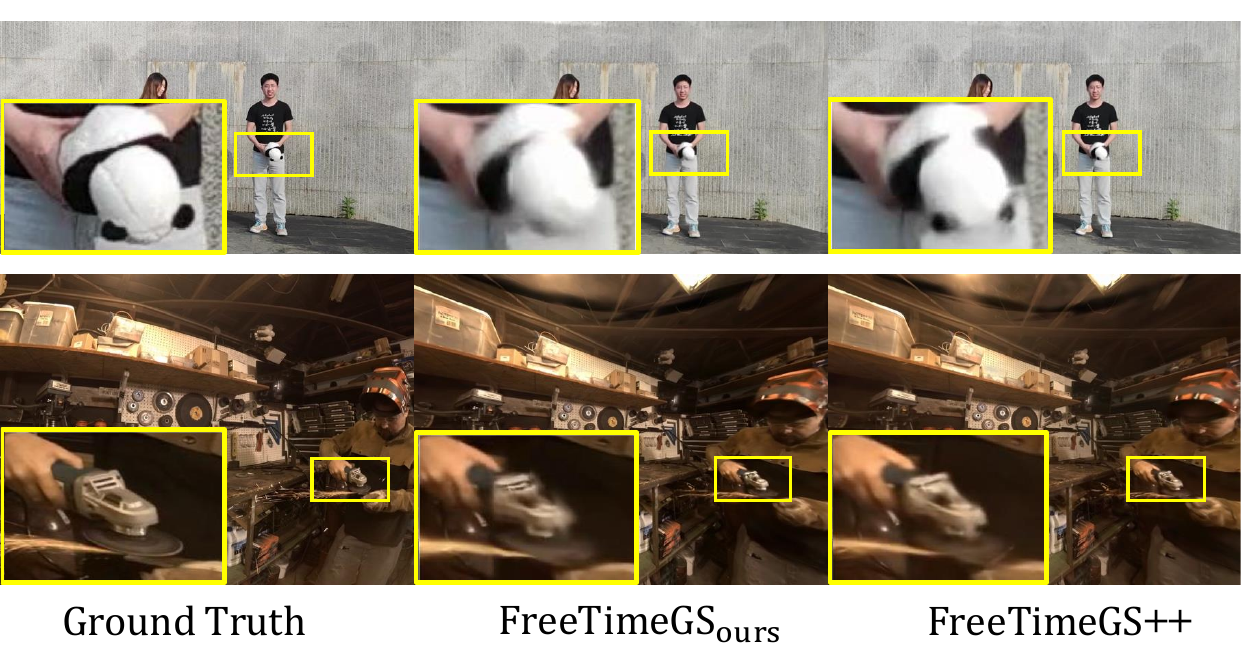}
  \caption{Qualitative comparison on ENeRF-Outdoor and Google Immersive.}
  \label{fig:outdoor-qualitative}
\end{figure}

\vspace{1mm}\noindent\textbf{Results on outdoor datasets.}
\label{subsec:additional-datasets}
We further evaluate \ftgspp on ENeRF-Outdoor and Google
Immersive to verify whether our method generalizes to outdoor environments.
As shown in \cref{tab:additional-datasets,fig:outdoor-qualitative}, \ftgspp
outperforms the reproduced baseline on both datasets, confirming that our
method remains effective in outdoor settings.
\begin{table}[t]
  \centering
  \caption{Outdoor dataset results.}
  \label{tab:additional-datasets}
  \vspace{-2mm}
  \small
  \setlength{\tabcolsep}{4pt}
  \begin{tabular}{llcc}
    \toprule
    Dataset & Method & PSNR $\uparrow$ & LPIPS$_\mathrm{Alex} \downarrow$ \\
    \midrule
    ENeRF-Outdoor & \ftgsours & 25.20 & 0.256 \\
    ENeRF-Outdoor & \ftgspp & \textbf{25.26} & \textbf{0.246} \\
    \midrule
    Google Immersive & \ftgsours & 22.29 & 0.094 \\
    Google Immersive & \ftgspp & \textbf{22.68} & \textbf{0.091} \\
    \bottomrule
  \end{tabular}
  \vspace{-1em}
\end{table}

\begin{figure}[t]
  \centering
  \includegraphics[width=\linewidth]{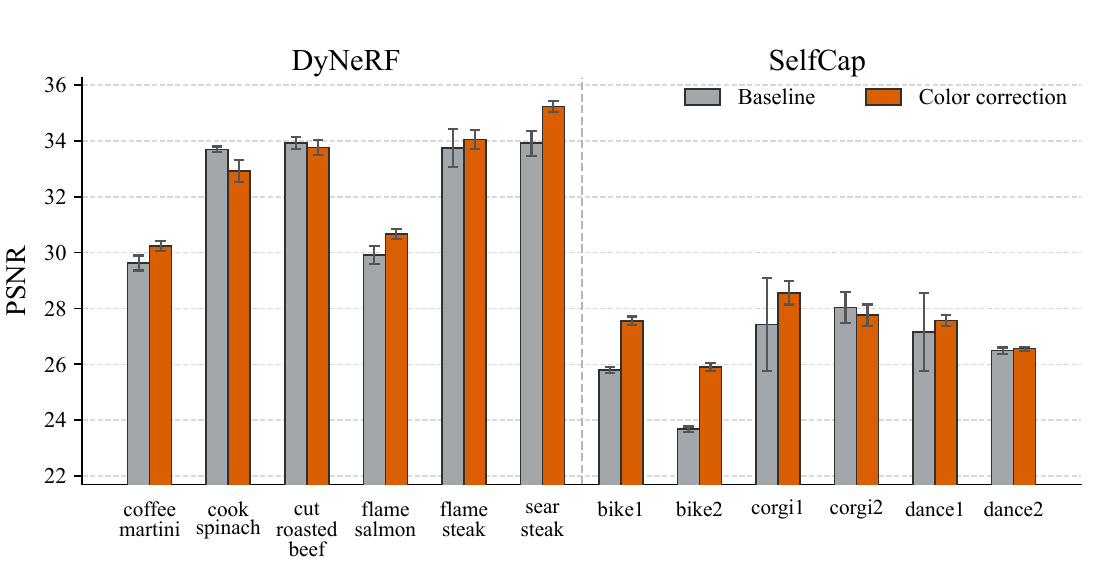}
  \vspace{-2em}
  \caption{Per-scene effect of color correction on PSNR and its
  run-to-run variance.}
  \label{fig:secret5_cc_metrics}
  \vspace{-2mm}
\end{figure}

\vspace{1mm}\noindent\textbf{Per-scene color correction.}
We further analyze the per-scene PSNR and standard deviation changes in
\Cref{fig:secret5_cc_metrics}.
Across most scenes, color correction improves PSNR and tends to reduce the
standard deviation.
This trend supports the repeatability analysis in \textbf{\textit{Secret 5}} (\Cref{subsec:evaluation_stability}), where single-run scores can mask the
run-to-run variation inherent in dynamic Gaussian reconstruction, and a
training-time stabilizer can mitigate this variation across scenes.

%% file: sec/7_limitations_conclusion.tex

\section{Conclusion}
\label{sec:conclusion}

In this paper, we have provided a systematic perspective on the
\textit{secrets} behind the 4D Gaussian Splatting performance.
By analyzing FreeTimeGS through our formalized baseline, we identified key
factors such as implicit temporal partitioning, photometric-motion decoupling,
spatiotemporal initialization, relocation behavior, and run-to-run repeatability.
Based on these observations, we propose \ftgspp, a principled framework
that replaces empirical heuristics with structurally consistent components.
The experimental results support the practical relevance of our analysis by showing that these factors can guide improvements over the
reproduced baseline.

\vspace{2mm}\noindent\textbf{Limitations and future work.} 
While our analysis focuses on FreeTimeGS as a representative baseline,
a systematic cross-method evaluation of all identified factors would further
strengthen the generality of our findings.
We include a preliminary check of color correction in the supplementary,
but a comprehensive study across diverse methods remains for future work.

%% file: sec/8_supplementary.tex
\renewcommand{\thefigure}{S\arabic{figure}}
\renewcommand{\thetable}{S\arabic{table}}
\setcounter{figure}{0}
\setcounter{table}{0}

\section{Experimental Details}
\label{sec:supp-exp-details}

We summarize the main experimental settings used throughout the paper.
Our released code includes the full configuration files and low-level implementation details.

\vspace{1mm}\noindent\textbf{Common setup.}
Unless otherwise specified, all experiments are trained for 30k iterations with
a maximum budget of 500k Gaussian primitives. We initialize 700k candidate
points per frame, use spherical harmonics of degree 3, and set the initial
opacity, spatial scale, and temporal duration to 0.5, 0.1, and 0.1,
respectively.

\vspace{1mm}\noindent\textbf{Initialization and relocation.}
RoMa matching uses the two nearest cameras for geometric initialization, and
initial velocities are estimated with $k=3$ nearest-neighbor correspondences.
Relocation starts at iteration 500, stops at iteration 25k, and is applied every
100 iterations. Primitives with opacity below $5\times10^{-3}$ are treated as
inactive.

\vspace{1mm}\noindent\textbf{FreeTimeGS++ components.}
For gated marginalization, we use $\gamma=20$, initialize gate logits to $-1$,
set the maximum duration to 1, and use gate regularization weight $10^{-3}$.
The color correction module is initialized as the identity transform,
regularized toward identity, and starts at iteration 3K with weight $10^{-3}$ by
default. UFM-guided initialization aggregates multi-view velocity estimates by
covisibility-weighted averaging and falls back to the KNN estimate when no valid
UFM estimate is available.

\vspace{1mm}\noindent\textbf{Repeated runs.}
For repeated-run experiments, run $r$ uses seed $S+r$ given a base seed $S$.
Unless otherwise specified, the default base seed is 0.

\section{LPIPS Evaluation Convention}
\label{sec:supp-lpips}

LPIPS~\cite{perceptual_metric} is sensitive to the input-range convention assumed
by the evaluation code.
The widely used Gaussian-splatting evaluation pipeline, inherited from
3DGS~\cite{3dgs}, normalizes inputs with statistics intended for the $[-1,1]$
range while passing images that remain in the $[0,1]$ range.
This mismatch makes the reported LPIPS systematically lower, and therefore more
favorable, than the value obtained under the mathematically intended $[-1,1]$
normalization, so numbers computed under the two conventions are not directly
comparable even when PSNR and SSIM agree.
Prior work has similarly cautioned that implementation and evaluation details can
materially affect reported Gaussian-splatting comparisons~\cite{bulo2024revising}.

To avoid this ambiguity, all LPIPS values in this paper follow a single, fixed
convention: rendered and ground-truth images are kept in the $[0,1]$ range and
passed to the LPIPS implementation consistently.
We adopt this prevailing Gaussian-splatting convention so that our numbers remain
directly comparable to FreeTimeGS~\cite{freetimegs} and other methods evaluated
the same way.
The same convention is used for our reproduced baseline \ftgsours, all \ftgspp
variants, and every additional check reported in the paper, and we do not mix
LPIPS values computed under different conventions.

\section{Per-Scene Quantitative Comparison}
\label{sec:supp-perscene}

We provide detailed per-scene quantitative results for the comparisons
summarized in the main paper.
Table~\ref{tab:reproducing} compares the numbers reported in
FreeTimeGS~\cite{freetimegs} with our reproduced baseline \ftgsours, and
Table~\ref{tab:final-model-improvement} compares \ftgsours with the improved
\ftgspp configuration.
Tables~\ref{tab:dynerf_perscene} and~\ref{tab:selfcap_perscene} present the
corresponding per-scene results on DyNeRF and SelfCap, respectively.
For prior methods, we use the per-scene numbers reported in
FreeTimeGS~\cite{freetimegs}.
\ftgsours achieves competitive per-scene performance under a 500k-Gaussian
budget, and \ftgspp further improves over this baseline across most scenes.

\begin{table*}[tp]
    \centering
    \caption{Quantitative comparison of view synthesis results on DyNeRF~\cite{dynerf} dataset. Green and yellow cell colors indicate
    \colorbox{colorfirst}{best} and \colorbox{colorsecond}{second} best results,
    respectively. For \ftgsours and \ftgspp, we report mean $\pm$ standard deviation over 6 independent runs.
    $^\dagger$ indicates results reported with no more than 500k primitives in
    the original paper~\cite{freetimegs}.}
    \renewcommand{\arraystretch}{1.0}
    \setlength{\tabcolsep}{4.5pt}
    \resizebox{0.9\textwidth}{!}{%
    \begin{tabular}{l cccccc c}
    \toprule
                    & Coffee Martini & Cook Spinach & Cut Roasted Beef & Flame Salmon &
                    Flame Steak & Sear Steak & Avg. \\
    \cmidrule(l){2-8}
                    & \multicolumn{7}{c}{PSNR$\uparrow$} \\
    \midrule
    Deformable-3DGS~\cite{4dgs-wu} & 27.34 & 32.46 & 32.90 & 29.20 & 32.51 & 32.49 & 31.15 \\
    4DGS~\cite{4dgs-yang} & 28.33 & 32.93 & 33.85 & 29.38 & 34.03 & 33.51 & 32.01 \\
    STGS~\cite{stg} & 28.61 & 33.18 & 33.52 & 29.48 & 33.64 & 33.89 & 32.05 \\
    STGS-Lite~\cite{stg} & 27.49 & 32.92 & 33.72 & 28.67 & 33.28 & 33.47 & 31.59 \\
    \ftgswang$^\dagger$~\cite{freetimegs} & \cellsecond{30.31} & \cellsecond{33.52} & 34.13 & \cellsecond{30.63} & \cellsecond{34.66} & \cellsecond{34.56} & \cellsecond{32.97} \\
    \midrule
    \textbf{\ftgsours} & 29.59 $\pm$ 0.40 & \cellfirst{33.65 $\pm$ 0.16} & \cellsecond{34.21 $\pm$ 0.25} & 30.10 $\pm$ 0.29 & 34.19 $\pm$ 0.33 & 33.96 $\pm$ 0.36 & 32.62 $\pm$ 0.14 \\
    \textbf{\ftgspp} & \cellfirst{30.62 $\pm$ 0.20} & 33.40 $\pm$ 0.31 & \cellfirst{34.50 $\pm$ 0.47} & \cellfirst{31.19 $\pm$ 0.08} & \cellfirst{34.91 $\pm$ 0.44} & \cellfirst{35.75 $\pm$ 0.24} & \cellfirst{33.40 $\pm$ 0.09} \\
    \midrule
                    & \multicolumn{7}{c}{DSSIM$_2\downarrow$} \\
    \midrule
    STGS & 0.0250 & 0.0113 & 0.0105 & 0.0224 & \cellsecond{0.0087} & 0.0085 & \cellsecond{0.014} \\
    STGS-Lite & 0.0270 & 0.0118 & 0.0112 & 0.0244 & 0.0097 & 0.0095 & 0.016 \\
    \ftgswang$^\dagger$ & 0.0214 & 0.0117 & 0.0110 & 0.0197 & 0.0095 & 0.0089 & \cellsecond{0.014} \\
    \midrule
    \textbf{\ftgsours} & \cellsecond{0.0177} & \cellfirst{0.0091} & \cellfirst{0.0081} & \cellsecond{0.0167} & \cellfirst{0.0074} & \cellfirst{0.0068} & \cellfirst{0.011} \\
    \textbf{\ftgspp} & \cellfirst{0.0170} & \cellsecond{0.0100} & \cellsecond{0.0090} & \cellfirst{0.0160} & 0.0090 & \cellsecond{0.0080} & \cellfirst{0.011} \\
    \midrule
                    & \multicolumn{7}{c}{LPIPS$_{Alex}\downarrow$} \\
    \midrule
    STGS & 0.069 & 0.037 & 0.036 & 0.063 & 0.029 & 0.030 & 0.044 \\
    STGS-Lite & 0.075 & 0.038 & 0.038 & 0.068 & 0.031 & 0.031 & 0.047 \\
    \ftgswang$^\dagger$ & 0.065 & 0.036 & 0.037 & 0.060 & 0.030 & \cellsecond{0.029} & \cellsecond{0.043} \\
    \midrule
    \textbf{\ftgsours} & \cellfirst{0.048} & \cellfirst{0.029} & \cellfirst{0.027} & \cellfirst{0.045} & \cellfirst{0.023} & \cellfirst{0.023} & \cellfirst{0.033} \\
    \textbf{\ftgspp} & \cellsecond{0.049} & \cellsecond{0.030} & \cellsecond{0.028} & \cellsecond{0.046} & \cellsecond{0.024} & \cellfirst{0.023} & \cellfirst{0.033} \\
    \bottomrule
    \end{tabular}}
\label{tab:dynerf_perscene}
\end{table*}

\begin{table*}[tp]
    \centering
    \caption{Quantitative comparison of view synthesis results on the
    \textit{SelfCap}~\cite{longvolcap}. Legend conventions follow
    Table~\ref{tab:dynerf_perscene}.}
    \renewcommand{\arraystretch}{1.0}
    \setlength{\tabcolsep}{4.5pt}
    \resizebox{0.9\textwidth}{!}{%
    \begin{tabular}{l cccccc c}
    \toprule
                    & dance\_1 & dance\_2 & corgi\_1 & corgi\_2 & bike\_1 & bike\_2 & Avg. \\
    \cmidrule(l){2-8}
                    & \multicolumn{7}{c}{PSNR$\uparrow$} \\
    \midrule
    Deformable-3DGS~\cite{4dgs-wu} & 26.82 & \cellsecond{26.54} & 24.15 & \cellsecond{28.04} & 24.30 & \cellsecond{25.25} & 25.95 \\
    4DGS-4DSH~\cite{4dgs-yang} & 26.49 & 24.99 & 27.08 & 27.56 & 24.61 & 24.42 & 25.98 \\
    4DGS~\cite{4dgs-yang} & 26.40 & 25.53 & 26.95 & 27.50 & 24.04 & 24.77 & 25.98 \\
    STGS~\cite{stg} & 25.26 & 23.74 & 27.55 & \cellfirst{28.52} & 21.46 & 22.09 & 24.97 \\
    \ftgswang$^\dagger$~\cite{freetimegs} & - & - & - & - & - & - & \cellfirst{27.27} \\
    \midrule
    \textbf{\ftgsours} & \cellfirst{27.61 $\pm$ 0.09} & \cellfirst{26.57 $\pm$ 0.13} & \cellsecond{28.23 $\pm$ 1.34} & 26.72 $\pm$ 2.78 & \cellsecond{25.88 $\pm$ 0.15} & 23.60 $\pm$ 0.15 & 26.43 $\pm$ 0.69 \\
    \textbf{\ftgspp} & \cellsecond{27.41 $\pm$ 0.08} & 26.36 $\pm$ 0.11 & \cellfirst{28.38 $\pm$ 0.46} & 27.44 $\pm$ 0.64 & \cellfirst{27.40 $\pm$ 0.22} & \cellfirst{25.73 $\pm$ 0.25} & \cellsecond{27.12 $\pm$ 0.08} \\
    \midrule
                    & \multicolumn{7}{c}{SSIM$_2\uparrow$} \\
    \midrule
    Deformable-3DGS & 0.940 & 0.936 & 0.900 & 0.902 & 0.920 & 0.924 & 0.926 \\
    4DGS-4DSH & 0.937 & 0.924 & 0.920 & 0.924 & 0.922 & 0.912 & 0.927 \\
    4DGS & 0.938 & 0.929 & 0.919 & 0.924 & 0.919 & 0.916 & 0.927 \\
    STGS & 0.933 & 0.916 & \cellsecond{0.933} & 0.934 & 0.822 & 0.827 & 0.905 \\
    \ftgswang$^\dagger$ & - & - & - & - & - & - & - \\
    \midrule
    \textbf{\ftgsours} & \cellfirst{0.960} & \cellfirst{0.953} & \cellfirst{0.949} & \cellsecond{0.936} & \cellfirst{0.949} & \cellsecond{0.931} & \cellsecond{0.946} \\
    \textbf{\ftgspp} & \cellsecond{0.957} & \cellsecond{0.951} & \cellfirst{0.949} & \cellfirst{0.943} & \cellsecond{0.948} & \cellfirst{0.933} & \cellfirst{0.947} \\
    \midrule
                    & \multicolumn{7}{c}{LPIPS$_{VGG}\downarrow$} \\
    \midrule
    Deformable-3DGS & 0.259 & 0.269 & 0.357 & 0.365 & 0.312 & 0.310 & 0.298 \\
    4DGS-4DSH & 0.246 & 0.269 & 0.212 & 0.210 & 0.256 & 0.275 & 0.239 \\
    4DGS & 0.244 & 0.257 & 0.211 & 0.212 & 0.259 & \cellsecond{0.269} & 0.237 \\
    STGS & \cellsecond{0.237} & 0.245 & 0.219 & 0.235 & 0.406 & 0.402 & 0.273 \\
    \ftgswang$^\dagger$ & - & - & - & - & - & - & \cellsecond{0.217} \\
    \midrule
    \textbf{\ftgsours} & \cellfirst{0.115} & \cellfirst{0.118} & \cellfirst{0.140} & \cellsecond{0.155} & \cellfirst{0.135} & \cellfirst{0.160} & \cellfirst{0.137} \\
    \textbf{\ftgspp} & \cellfirst{0.115} & \cellsecond{0.119} & \cellsecond{0.141} & \cellfirst{0.151} & \cellsecond{0.138} & \cellfirst{0.160} & \cellfirst{0.137} \\
    \bottomrule
    \end{tabular}}
\label{tab:selfcap_perscene}
\end{table*}

\section{Color Correction on Other Baseline}
\label{sec:supp-cc-stgs}

We further examine whether the training-time color correction (CC) module behaves
consistently outside the reproduced FreeTimeGS baseline.
To this end, we apply the same CC module to STGS~\cite{stg}, a representative
dynamic Gaussian splatting method, on DyNeRF.

This experiment is intended as a within-method comparison between the original
STGS baseline and its CC-augmented variant.
Both variants are evaluated under the same primitive budget, training length,
frame sampling setting, and repeated-run protocol.
We use a 500k primitive budget and repeat each setting over 6 independent runs.
Although STGS is commonly trained for 30k iterations in its default setting, we
fix the training length to 15k iterations for this repeated-run analysis due to
computational cost.
For frame sampling, we follow the default setting of the public repository and
use 50 frames.

\begin{table*}[t]
  \centering
  \caption{Color correction check for STGS on DyNeRF.
  Each PSNR entry reports mean $\pm$ standard deviation over 6 independent runs.}
  \label{tab:cc-stgs}

  \begin{minipage}[t]{0.60\textwidth}
    \centering
    \vspace{2mm}
    \small
    \resizebox{\linewidth}{!}{%
    \begin{tabular}{lcccccc}
      \toprule
      \multirow{2}{*}{Scene} &
      \multicolumn{2}{c}{PSNR $\uparrow$} &
      \multicolumn{2}{c}{LPIPS$_\mathrm{Alex}$ $\downarrow$} &
      \multicolumn{2}{c}{SSIM$_2$ $\uparrow$} \\
      \cmidrule(lr){2-3}
      \cmidrule(lr){4-5}
      \cmidrule(l){6-7}
      & Baseline & +\CC & Baseline & +\CC & Baseline & +\CC \\
      \midrule
      coffee\_martini    & 28.42 $\pm$ 0.22 & 28.60 $\pm$ 0.17 & 0.074 & 0.075 & 0.908 & 0.909 \\
      cook\_spinach      & 33.22 $\pm$ 0.18 & 33.51 $\pm$ 0.18 & 0.038 & 0.037 & 0.953 & 0.953 \\
      cut\_roasted\_beef & 32.94 $\pm$ 0.41 & 34.83 $\pm$ 0.21 & 0.039 & 0.037 & 0.954 & 0.956 \\
      flame\_salmon      & 28.88 $\pm$ 0.20 & 29.59 $\pm$ 0.09 & 0.065 & 0.064 & 0.918 & 0.919 \\
      flame\_steak       & 33.56 $\pm$ 0.44 & 35.74 $\pm$ 0.20 & 0.032 & 0.030 & 0.961 & 0.960 \\
      sear\_steak        & 33.62 $\pm$ 0.51 & 36.07 $\pm$ 0.28 & 0.032 & 0.030 & 0.961 & 0.961 \\
      \midrule
      \textbf{\textit{Avg.}} &
      \textbf{31.77 $\pm$ 0.33} & \textbf{33.06 $\pm$ 0.19} &
      \textbf{0.047} & \textbf{0.045} &
      \textbf{0.942} & \textbf{0.943} \\
      \bottomrule
    \end{tabular}}
  \end{minipage}
  \hfill
  \begin{minipage}[t]{0.38\textwidth}
    \centering
    \vspace{0mm}
    \includegraphics[width=\linewidth]{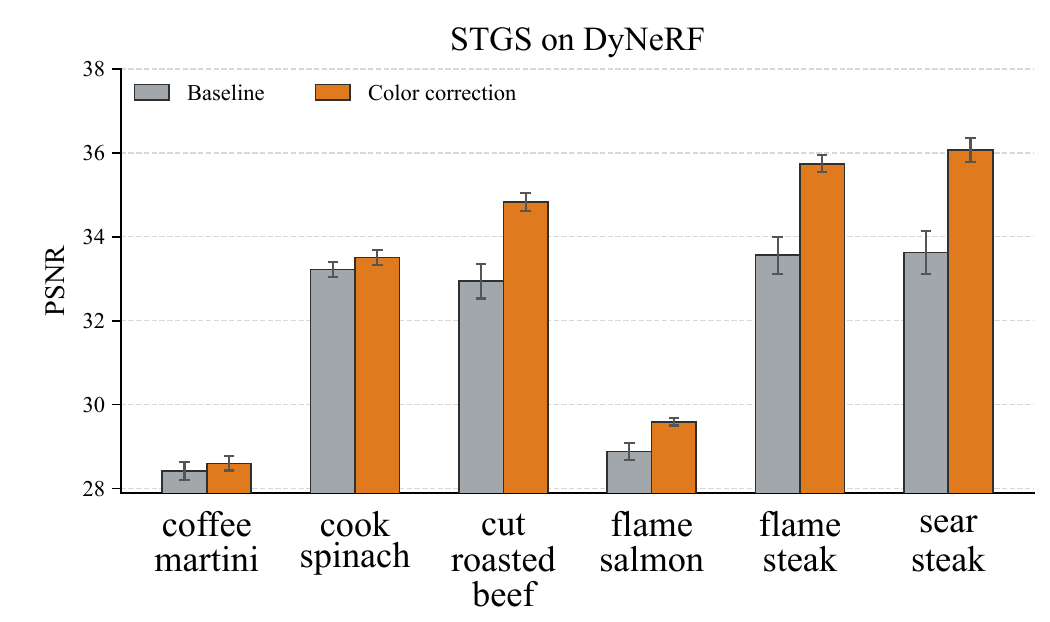}
  \end{minipage}
\end{table*}

As shown in \cref{tab:cc-stgs}, adding CC to STGS generally reduces the PSNR
standard deviation on DyNeRF.
This indicates that the training-time CC module also behaves as a stabilizing
component in another dynamic Gaussian splatting pipeline outside our reproduced
FreeTimeGS baseline.
Consistent with the trend observed in our baseline, this stabilization is also
accompanied by improved PSNR on the evaluated scenes.

\section{Effective Lifetime of Temporal Gaussians}
\label{sec:supp-lifetime}

This section derives the closed-form lifetime summarized in
\cref{subsec:representation}.
Let $T$ denote the temporal span of the training sequence.
As defined in \cref{subsec:base-representation}, each Gaussian modulates its base
opacity $o$ by the Gaussian-shaped activation 
\[
  a(t)=\exp[-\tfrac{1}{2}((t-\mu_t)/s)^2],
\]
with temporal center $\mu_t$ and
temporal scale $s$.
We treat a primitive as visible at time $t$ when its rendered opacity $o\,a(t)$
exceeds a visibility threshold $\theta$.
Solving $o\,a(t)=\theta$ for $t$ gives a symmetric visible window around $\mu_t$,
whose width defines the \textit{effective lifetime}
\begin{equation}
\Delta t(\theta)
=
2 s\sqrt{2\log(o/\theta)}
=
\frac{d}{3}\sqrt{2\log(o/\theta)},
\end{equation}
for $o>\theta$ and zero otherwise, where $d=6s$ is the nominal duration.
The effective lifetime thus depends jointly on the duration $d$ and the base
opacity $o$ rather than on duration alone, so two primitives with the same
duration can remain visible for different lengths of time.
For the distribution in \cref{fig:secret1_partitioning}(a), we compute
$\Delta t(\theta)/T$ for every primitive after convergence.

\vspace{2mm}\section{Computational Cost of UFM Initialization}
\label{sec:supp_ufm_cost}

Table~\ref{tab:ufm_time_cost} reports the average runtime of
UFM~\cite{ufm}-guided temporal initialization.
We separate the one-time flow precomputation cost from the subsequent
initialization cost using the cached flows.
With the flow cache, the additional initialization cost remains on the order of
tens of seconds over the baseline k-NN initialization.
The larger cost without the cache mainly comes from running optical flow over
the sampled frames and cameras: DyNeRF uses 30 sampled frames with 18 cameras,
whereas SelfCap uses 60 sampled frames with 24 cameras.
All initialization benchmarks are measured on an AMD EPYC 9554 CPU using a
single process capped at 16 threads, and flow precomputation is accelerated by a
single NVIDIA RTX PRO 6000 Blackwell MAX-Q Workstation Edition GPU.

\begin{table}[t]
  \centering
  \caption{Average runtime of UFM-guided temporal initialization.}
  \label{tab:ufm_time_cost}
  \small
  \begin{tabular*}{\linewidth}{@{\extracolsep{\fill}}lccccc@{}}
    \toprule
    \multirow{2}{*}{Dataset} &
    \multirow{2}{*}{\shortstack{Flow\\frames}} &
    \multirow{2}{*}{Cams} &
    \multirow{2}{*}{\shortstack{Baseline\\Init}} &
    \multicolumn{2}{c}{UFM-Guided Init} \\
    \cmidrule(l){5-6}
    & & & & w/ cache & w/o cache \\
    \midrule
    DyNeRF & 30 & 18 & 17.5s & 44.1s & 741.3s \\
    SelfCap & 60 & 24 & 20.7s & 80.8s & 1707.8s \\
    \bottomrule
  \end{tabular*}
\end{table}

\section{Visualization of UFM-Guided Initialization}
\label{sec:supp_ufm_vis}

The photometric--motion decoupling observed in \cref{subsec:representation}
shows that rendering metrics alone do not reveal the internal motion behavior of
the learned Gaussians.
To further inspect the effect of UFM-guided initialization in
\cref{subsec:improved-ufm}, we visualize per-Gaussian motion as velocity maps.

\begin{figure}[t]
  \centering
  \vspace{2mm}
  \includegraphics[width=\linewidth]{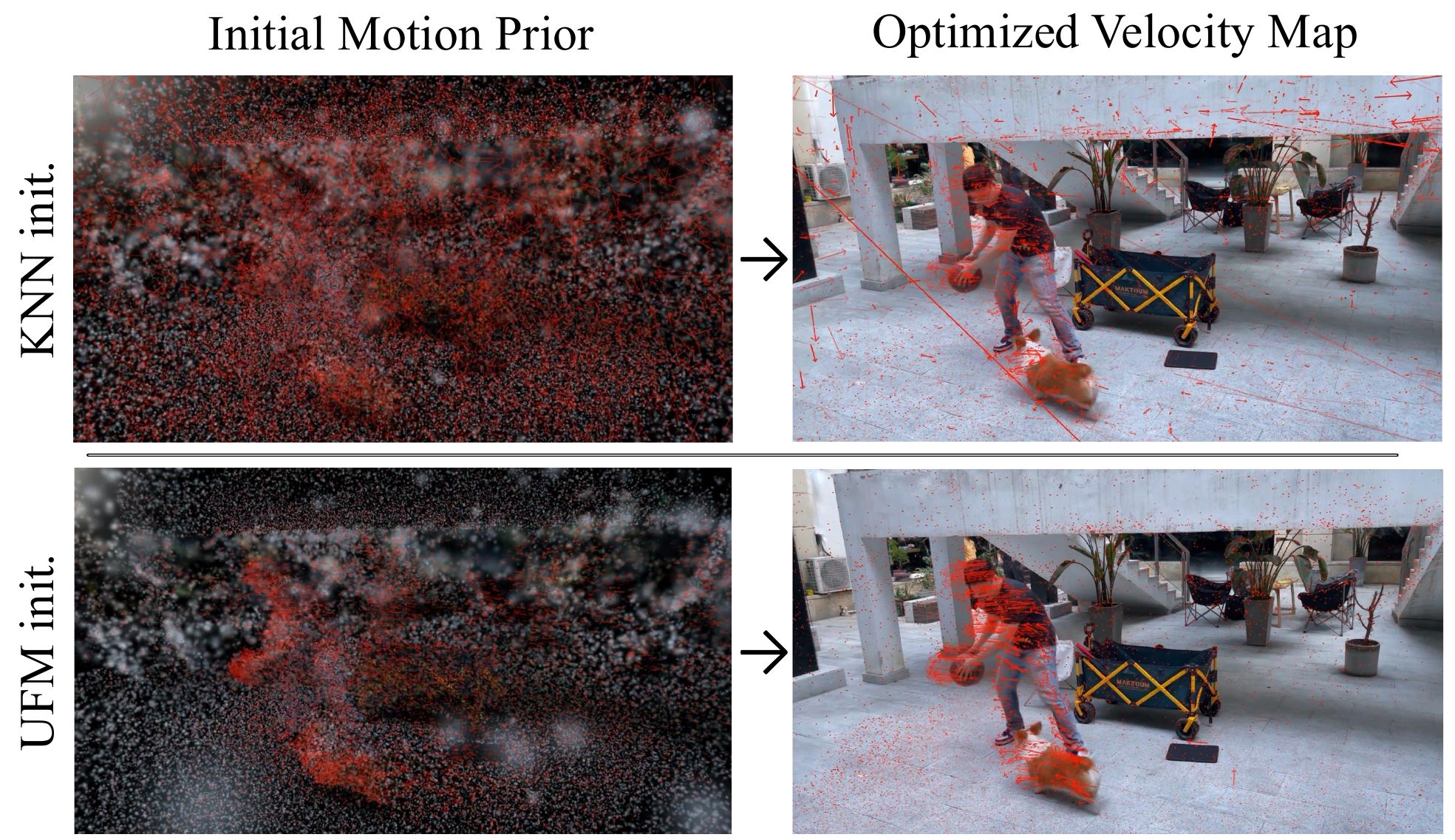}
  \caption{UFM-guided initialization on \textit{Corgi2} from SelfCap.}
  \label{fig:velocity_comparison}
\end{figure}

\Cref{fig:velocity_comparison} compares KNN-based and UFM-guided initialization
on the SelfCap \textit{Corgi2} scene at the same timestamp.
We show the velocity maps before optimization and after optimization.
Compared with the KNN-based initialization, UFM-guided initialization provides a
cleaner and more spatially coherent motion estimate at initialization, and this
structure remains more coherent after optimization.
This suggests that the UFM prior provides a more motion-aware starting
point, complementing the rendering-based evaluation in the main paper.

